%% file: main.tex
\documentclass[10pt, conference]{format/ieeeconf}

\IEEEoverridecommandlockouts    
\usepackage{glossaries}
\input{glossary}

\usepackage{tabularx}
\usepackage{multirow, multicol}
\usepackage{array}
\usepackage{tabu}

\newcolumntype{P}[1]{>{\centering\arraybackslash}p{#1}}
\newcolumntype{M}[1]{>{\centering\arraybackslash}m{#1}}
\usepackage{booktabs}
\newcolumntype{N}{>{\centering\arraybackslash}m{.5in}}
\newcolumntype{G}{>{\centering\arraybackslash}m{2in}}

\let\labelindent\relax

\usepackage[table,usenames,dvipsnames]{xcolor}      
\usepackage{extarrows}                              
\usepackage{enumitem}
\usepackage{wrapfig}

\usepackage{amsmath,amssymb,amsfonts,amsthm,dsfont, bm} 
\usepackage{mathtools}
\usepackage{algorithm,algorithmicx,listings}        
\usepackage[noend]{algpseudocode}			              
\makeatletter
\def\BState{\State\hskip-\ALG@thistlm}
\makeatother
\DeclarePairedDelimiter\abs{\lvert}{\rvert}%
\DeclarePairedDelimiter\norm{\lVert}{\rVert}%
\makeatletter
\let\oldabs\abs
\def\abs{\@ifstar{\oldabs}{\oldabs*}}
\let\oldnorm\norm
\def\norm{\@ifstar{\oldnorm}{\oldnorm*}}
\makeatother

\usepackage{graphicx}
\usepackage{subcaption}
\usepackage{textcomp}
\usepackage[font=footnotesize]{caption}
\usepackage{subcaption}
\usepackage[breaklinks=true, colorlinks, bookmarks=true, citecolor=Black, urlcolor=Violet,linkcolor=Black]{hyperref}

\usepackage[normalem]{ulem}
\usepackage[utf8]{inputenc}
\usepackage[english]{babel}
\usepackage{hyperref}
\hypersetup{
    colorlinks=true,
    linkcolor=blue,
    filecolor=magenta,      
    urlcolor=cyan,
}

\DeclareMathAlphabet\mathbfcal{OMS}{cmsy}{b}{n}

\usepackage[space, compress, sort]{cite}

\newtheorem*{assumption*}{Assumption}

\newtheorem*{problem*}{Problem}

\usepackage{lipsum}
\usepackage[capitalise]{cleveref}

\begin{document}

\title{Learning to Explore Indoor Environments using \\ Autonomous Micro Aerial Vehicles}

\author{Yuezhan Tao$^{1}$, Eran Iceland$^{2}$, Beiming Li$^{1}$, Elchanan Zwecher$^{2}$, Uri Heinemann$^{2}$, \\
Avraham Cohen$^{3}$, Amir Avni$^{3}$, Oren Gal$^{3}$, Ariel Barel$^{3}$ and Vijay Kumar$^{1}$ %
\thanks{$^{1}$GRASP Laboratory, University of Pennsylvania, Philadelphia, PA, 19104, USA {\tt\small\{yztao, beimingl, kumar\} @seas.upenn.edu}.} %
\thanks{$^{2}$Hebrew University of Jerusalem, Jerusalem, Israel {\tt\small\{eran.iceland,elchanan4567,urihei\}@gmail.com.}}%
\thanks{$^{3}$Technion - Israel Institute of Technology, Haifa, Israel {\tt\small \{avico@, orengal@alumni., arielba@\}technion.ac.il, amiravni83@gmail.com.}}%
\thanks{We gratefully acknowledge the support of 
ARL DCIST CRA W911NF-17-2-0181, 
NSF Grants CCR-2112665, 
EEC-1941529, 
and ONR grant N00014-20-1-2822. }%
}
\maketitle

\begin{abstract}

In this paper, we address the challenge of exploring unknown indoor aerial environments using autonomous aerial robots with \gls{swap} constraints. The \gls{swap} constraints induce limits on mission time requiring efficiency in exploration. We present a novel exploration framework that uses \gls{dl} to predict the most likely indoor map given the previous observations, and \gls{drl} for exploration, designed to run on modern SWaP constraints neural processors. The DL-based map predictor provides a prediction of the occupancy of the unseen environment while the DRL-based planner determines the best navigation goals that can be safely reached to provide the most information. The two modules are tightly coupled and run onboard allowing the vehicle to safely map an unknown environment. Extensive experimental and simulation results show that our approach surpasses state-of-the-art methods by 50-60\% in efficiency, which we measure by the fraction of the explored space as a function of the length of the trajectory traveled.

\end{abstract}

    \input{tex/Introduction}

    \input{tex/RelatedWork}

\input{tex/Framework_and_Algo_Design}

    \input{tex/Sim_and_Exp}

    \input{tex/Conclusions}

\bibliography{literature}

\end{document}

%% file: glossary.tex
\newacronym{mav}{MAV}{Micro Aerial Vehicles}
\newacronym{uav}{UAV}{Unmanned Aerial Vehicle}
\newacronym{ovc}{OVC}{Open Vision Computer}
\newacronym{lidar}{LiDAR}{Light Detection and Ranging}
\newacronym{vio}{VIO}{visual-inertial odometry}
\newacronym{gpgpu}{GPGPU}{General-Purpose Graphics Processing Unit}
\newacronym{ugv}{UGV}{Unmanned Ground Vehicle}
\newacronym{uwb}{UWB}{Ultra Wideband}
\newacronym{svm}{SVM}{Support Vector Machine}
\newacronym{fcn}{FCN}{Fully Convolutional Network}
\newacronym{cnn}{CNN}{Convolutional Neural Network}
\newacronym{loam}{LOAM}{LiDAR Odometry and Mapping}
\newacronym{sloam}{SLOAM}{Semantic LiDAR Odometry and Mapping}
\newacronym{slam}{SLAM}{Simultaneous Localization and Mapping}
\newacronym{iot4ag}{IoT4Ag}{NSF Engineering Research Center for the Internet of Things for Precision Agriculture}
\newacronym{grasp-lab}{GRASP Lab}{the General Robotics, Automation, Sensing and Perception Laboratory}
\newacronym{jps}{JPS}{Jump Point Search}
\newacronym{ukf}{UKF}{Unscented Kalman Filter}
\newacronym{sam}{SAM}{Smoothing and Mapping}
\newacronym{icp}{ICP}{Iterative Closest Point}
\newacronym{imu}{IMU}{Inertial Measurement Unit}
\newacronym{tsdf}{TSDF}{Truncated Signed Distance Field}
\newacronym{esdf}{ESDF}{Euclidean Signed Distance Field}
\newacronym{sdf}{SDF}{Signed Distance Field}
\newacronym{rrt}{RRT}{Rapidly Exploring Random Tree}
\newacronym{fpv}{FPV}{First-person View}
\newacronym{dnn}{DNN}{Deep Neural Network}
\newacronym{igpred}{IGPred}{Information Gain Prediction}
\newacronym{csqmi}{CSQMI}{Cauchy-Schwarz Quadratic Mutual Information}
\newacronym{nbv}{NBV}{Next Best View}
\newacronym{vae}{VAE}{Variational Autoencoder}
\newacronym{tsp}{TSP}{Traveling Salesman Problem}
\newacronym{bcsm}{BCSM}{Behavior Control State Machine}
\newacronym{pca}{PCA}{Principal Component Analysis}
\newacronym{aspp}{ASPP}{Atrous Spatial Pyramid Pooling}
\newacronym{swap}{SWaP}{Size Weight and Power}
\newacronym{soi}{SoI}{Semantic Object of Interest}
\newacronym{aoi}{AoI}{Area of Interest}
\newacronym{drl}{DRL}{Deep Reinforcement Learning}
\newacronym{dl}{DL}{Deep Learning}
\newacronym{fov}{FoV}{Field of View}

%% file: tex/Introduction.tex
\section{Introduction}
\label{sec:intro}

Autonomous exploration entails robots navigating in an unknown environment to construct a detailed map. Its practical applications span diverse scenarios, such as search and rescue missions and automated area inspections. \gls{mav}, particularly quadrotors, have gained growing popularity for addressing such tasks, because of their capability to maneuver over low-height obstacles and accomplish exploration tasks with great agility and safety.

The problem of autonomous exploration has been widely studied. Over the past decade, various frameworks have been proposed to enhance performance by reducing the mapping uncertainty~\cite{KelseyIG, LukasIG, charrow2015CSQMI}, total time~\cite{zhou2021fuel}, or travel distance~\cite{mixed2017FrontierTSP, zhou2021fuel} when achieving a certain percentage of space coverage. A specific area of research concentrates on the incorporation of map predictors to empower traditional exploration frameworks. The prediction module utilizes prior statistics to estimate unseen portions of the environment, thereby enabling improved evaluation of information gain and uncertainty~\cite{2dMapPred2019, uncertaintyPrediction2019, SEER}. This, in turn, boosts the performance of the underlying exploration strategy. Moreover, recent success in applying Deep Reinforcement Learning (DRL) techniques to autonomous exploration problems presents promising results comparable to classical frontier-based methods~\cite{drlexploration2019, botteghi2021curiosity, betterRL2019, 9811861}. 

\begin{figure}[t!]
    \centering
    \includegraphics[width=0.95\columnwidth]{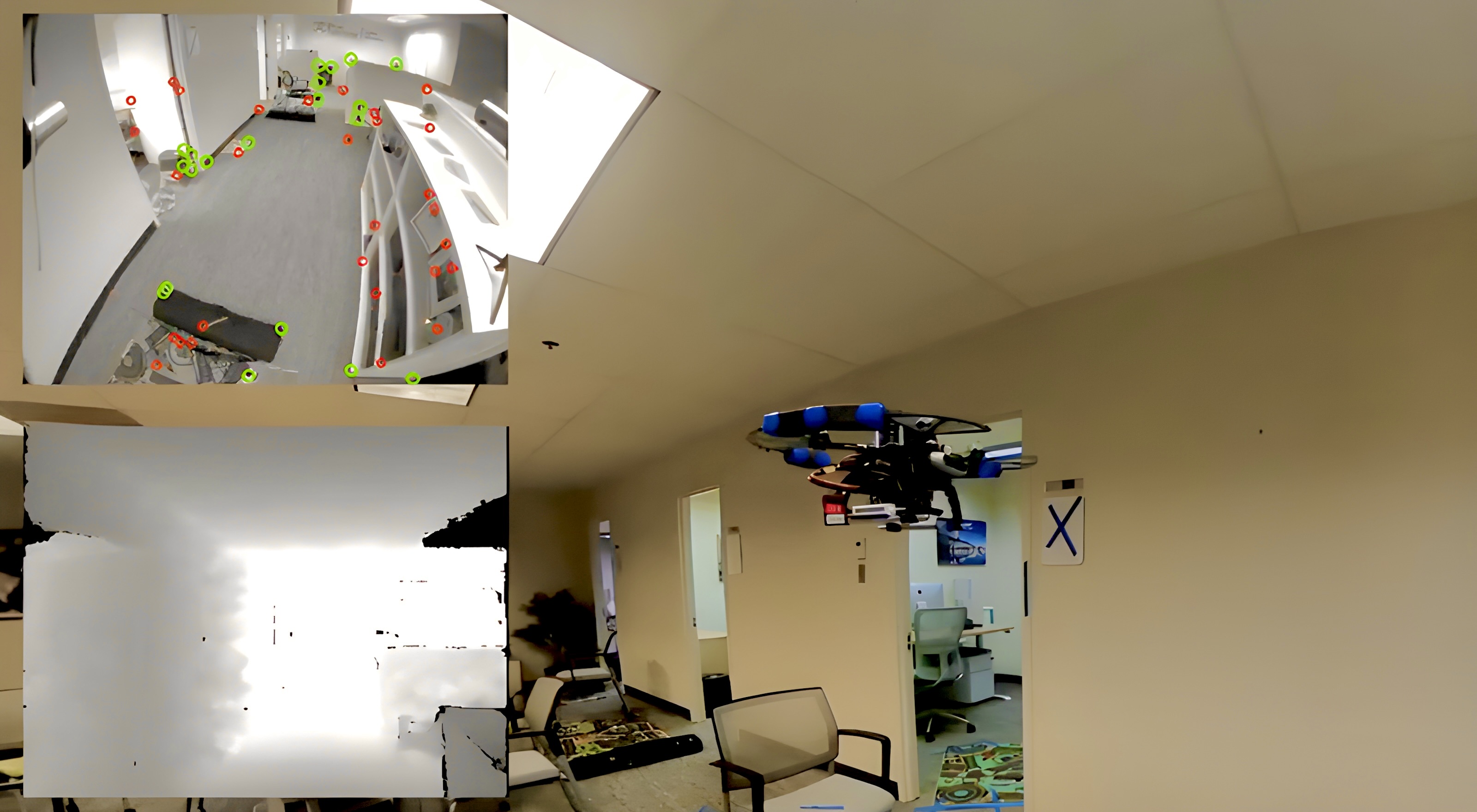}
    \caption{Our Falcon 250 Platform~\cite{SEER} exploring an office building. Falcon 250 is equipped with an Intel Realsense D435i camera, a VOXL board~\cite{voxl}, a Pixhawk 4 Mini flight controller, and an Intel NUC 10 onboard computer. The top left window displays the grayscale image with VIO features overlaid, where green/red circles mark high/low-quality features. The bottom left window displays the raw depth image from the RealSense camera.}
    \label{fig:platform}
    \vspace{-.8cm}
\end{figure}

Recent work proposed exploration frameworks with hierarchical algorithms and well-crafted utility functions. The results indicate significant improvements in exploration efficiency, but numerous tunable parameters are introduced and fine-tuning is required when switching task environments. While the emerging Deep Learning (DL)-based map predictor offers a way to enhance the utility estimation with predicted information, existing work such as~\cite{2dMapPred2019} still selects exploration goals from observed maps rather than predicted ones. Further, even though \gls{drl}-based approaches yield results that are comparable to those obtained from classical methods, it remains unclear if they can achieve similar results as state-of-the-art exploration frameworks.

In this paper, we study the fusion of the map prediction module and \gls{drl}-based exploration. \gls{drl}-based exploration enhances the framework's generalizability by eliminating the necessity of manually designing and fine-tuning utility functions. We propose a novel way to incorporate the map predictor's accuracy into the \gls{drl} training process, which establishes a tight coupling between these two modules. The framework's capability to formulate exploration decisions using both the observed map and the predicted map simultaneously equips it with the potential to surpass state-of-the-art methods in terms of efficiency, as indicated by the ratio of space coverage to trajectory length. In summary, our contributions are:
\begin{itemize}
    \item We propose a novel autonomous exploration framework that tightly couples a \gls{dl}-based map predictor and a \gls{drl}-based exploration planner for \gls{swap} constrained platforms.
    \item We validate the effectiveness of the proposed framework in various simulated environments and demonstrate its superiority over existing algorithms. 
    \item We deploy the proposed framework on \gls{swap} constrained MAV platforms and conduct real-world experiments where all sensing and processing are on board and no external infrastructure is needed.
\end{itemize}

%% file: tex/RelatedWork.tex
\section{Related Work}
\label{sec:related work}

\subsection{Efficient Autonomous Exploration}
Various methods have been proposed to solve the problem of autonomous exploration. The frontier-based approach that focuses on expanding the map into the unknown region was introduced in \cite{Frontier}, and extended into 3-D scenarios \cite{shen2012indoor}. By continuously navigating to specific regions to reduce the uncertainty of the map, the information-based approach serves as another heuristic for exploration. Different sampling methods and information evaluation algorithms are proposed in recent works \cite{charrow2015CSQMI, LukasIG , KelseyIG}\cite{alexis2020MP, NBVexplore2016}. However, since information can also be obtained by revisiting the explored region, the information-based approach leads to higher map quality but poorer exploration efficiency. To further improve exploration efficiency, refined utility functions have been proposed~\cite{topo2020indoor, ramon2019semanticIndoor}, along with hierarchical exploration strategies. Exploration efficiency is improved by navigating toward frontiers while maximizing information gain through local trajectory planning \cite{charrow2015information, mixed2019SampleVP}. In \cite{zhou2021fuel, mixed2017FrontierTSP}, \gls{tsp} is used to plan global paths that travel through frontiers to maximize information gain or coverage. Nonetheless, hierarchical approaches typically involve numerous tunable parameters to control waypoint sampling and utility functions, which can lead to unstable performance when applied to different environments. Following the recent success of deep reinforcement learning, DRL-based approaches gained popularity in exploration tasks \cite{barratt2017active, drlexploration2019, OccUnvertaintyPrediction2022, botteghi2021curiosity}. The results in \cite{drlexploration2019, botteghi2021curiosity} demonstrate that DRL can achieve performance similar to frontier-based approaches. In \cite{betterRL2019}, a frontier selection policy is learned and the resulting planner outperforms frontier-based methods. However, it is still unclear whether DRL-based methods could yield better results than the state-of-the-art exploration method.

\subsection{Map Prediction for Exploration}
Prediction of space occupancy in unknown environments can significantly contribute to navigation and exploration tasks. This predictive process can be generally divided into database-driven or deep learning-based approaches. In the former, historical data is employed to deduce unexplored information, as evidenced in \cite{mapRetrival2015}. With the ability to learn from and adapt to extensive historical data, the usage of deep learning for map prediction has gained increasing popularity in recent times. The focus of most existing research lies in leveraging structural information to predict occupancy and estimate the information gain for each candidate exploration goal. In \cite{2dMapPred2019, uncertaintyPrediction2019, SEER}, the generative map prediction model aids in the estimation of potential information or uncertainty. Egocentric 2-D occupancy is predicted in~\cite{OccUnvertaintyPrediction2022} to help the exploration and navigation tasks. Although previous work does introduce map prediction as an additional module to the traditional exploration framework, the predicted outputs are not fully utilized, as the navigation goals are still determined on observed maps.

Building on the idea of our previous work~\cite{9811861}, we propose a novel exploration framework that fuses a DL-based map predictor with a DRL-based exploration planner. In particular, we design a reward function based on the prediction's completeness and accuracy, which equips the planner with the ability to select navigation goals that promote a more certain and thorough prediction. Furthermore, the DRL-based planner takes in observations composed of both observed maps and predicted maps. While the former aids in collision avoidance while traversing cluttered and noisy environments, the latter provides prescient vision for selecting next actions. Importantly, we also develop a software architecture that allows for all modules to run on board on an off-the-shelf NUC 10 computer.

%% file: tex/Framework_and_Algo_Design.tex
\section{System Overview}
\begin{figure}[t!]
    \centering
    \includegraphics[width=0.9\columnwidth]{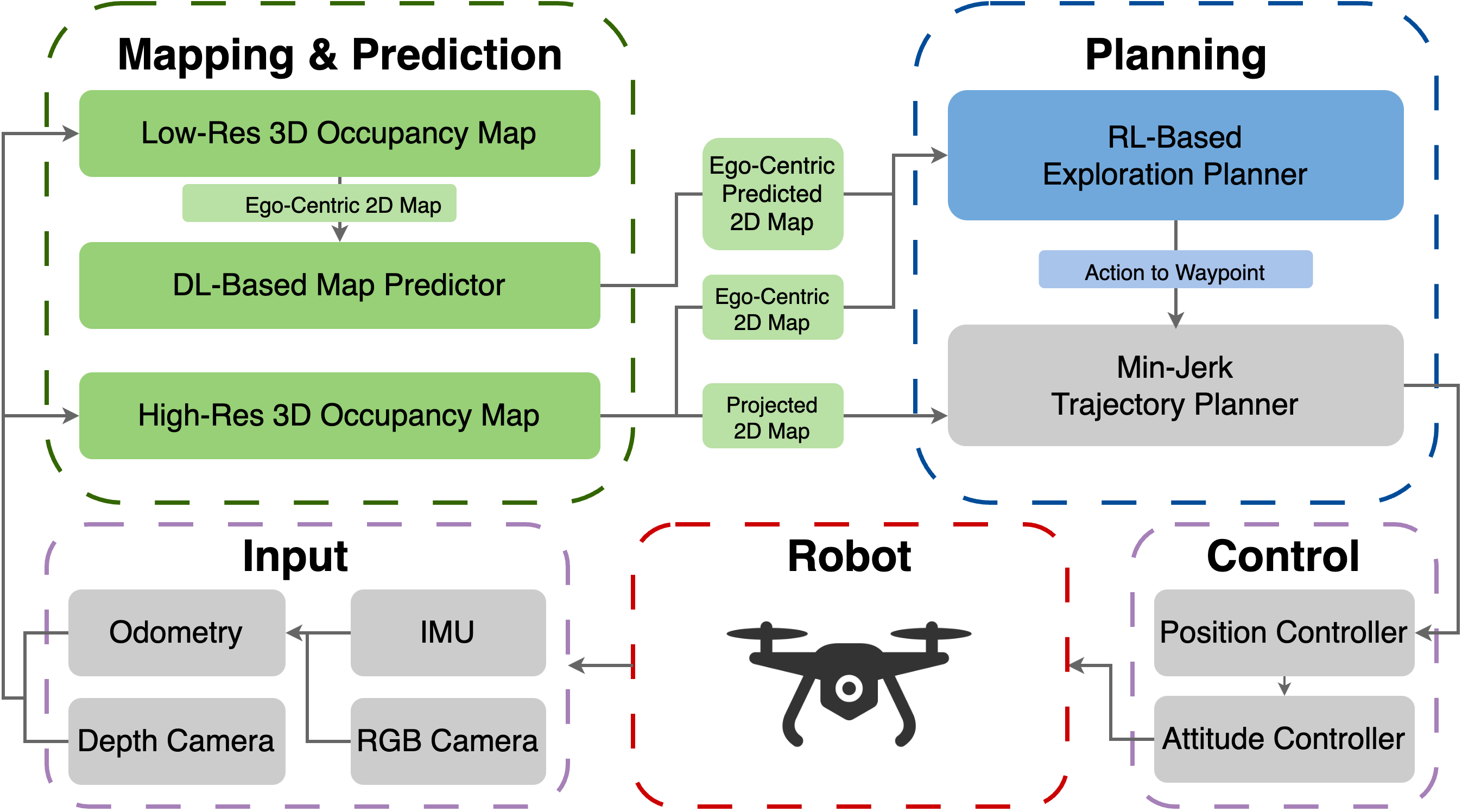}
    \caption{Proposed system architecture. The robot uses onboard color and depth cameras along with an IMU for sensing. The software architecture consists of two main modules: the mapping and DL-based map prediction module, and the planner module which includes a DRL-based exploration planner and a safe trajectory planner.}
    \label{fig:sys_diagram}
    \vspace{-.8cm}
\end{figure}
As illustrated in Fig.~\ref{fig:sys_diagram}, the proposed aerial system contains two main modules. The mapping and prediction module (Sect.~\ref{sec:map_prediction}) takes the raw depth image and estimated odometry as input to construct high-resolution and low-resolution 3-D occupancy maps. 2-D occupancy maps are generated from low-resolution 3-D occupancy maps and are used by the DL-based map predictor. The exploration planning module (Sect.~\ref{sec:rl_planner}) takes the 2-D occupancy grid projected from the high-resolution occupancy map, the predicted map, together with its previous action, to generate exploration actions, which are then converted to waypoints for planning minimum jerk trajectories. Network architectures for both DL and DRL modules, are optimized to fulfill both the desired mapping and exploration performance as well as memory and TOPS requirements for SWaP constraints neural processors.

\section{Mapping and Prediction}
\label{sec:map_prediction}
In this section, we introduce a mapping and prediction module that maintains a hierarchical map representation and predicts occupancy information over the unknown region.

\subsection{Hierarchical Mapping}
Maintaining a two-level hierarchical map presentation is beneficial for autonomous exploration tasks. The high-resolution map provides details about the environment for planning and collision avoidance, while the low-resolution map is favorable for map prediction and decision-making processes. In our proposed software system, the DRL-based exploration planner utilizes the two-level hierarchical map to make decisions for the next action. 

As shown in Fig.~\ref{fig:sys_diagram}, when the robot navigates through the environment, depth images associated with the estimated odometry are obtained and passed into the mapping and prediction module. When depth images and odometry data come in, log-odds-based map updates are conducted to construct high-resolution and low-resolution 3-D voxel maps, $V_{high}$ and $V_{low}$, respectively.
Then, $V_{high}$ and $V_{low}$ are projected along the z-axis to generate 2-D occupancy grid maps, $M_{high}$ and $M_{low}$. $M_{low}$ is padded and shifted to generate the ego-centric map, $M_{low}^{ego}$, which is then passed into the DL-based map predictor to generate, $M_{pred}^{ego}$, the ego-centric map containing predicted occupancy information. Dynamic thresholding, which will be introduced in Sect.~\ref{subsec:predictor}, is used for converting $M_{pred}^{ego}$ into $M_{thres}^{ego}$. $M_{thres}^{ego}$ is then cropped to produce ego-centric state $S_{thres}^{ego}$, which is used by the DRL-based planner for exploration planning.  
In parallel, $M_{high}$ is padded and shifted to produce ego-centric high-resolution state $S_{high}^{ego}$, serving as the second input to the DRL-based planner.

\subsection{Dataset}
\begin{figure}[t!]
    \centering
    \includegraphics[trim=0 0 0 0, clip, width=0.45\textwidth]{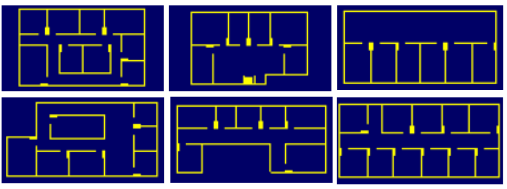}
    \caption{Sample building maps from training set}
    \label{fig:floorplans}
    \vspace{-.7cm}
\end{figure}

To train our robot, we created a dataset comprising of 50,000 2-D floor plans with a maximum size of 21m x 11m. All walls in our data set are perpendicular. We set the minimum door width to 80 cm and assume that all doors are open. While most of the rooms are situated adjacent to buildings' perimeters, we also allow for internal rooms that are not connected to external walls. The buildings' perimeters are not necessarily convex. We introduced primitives to represent furniture within the buildings. Low-height furniture like chairs and tables are not considered part of the floor plans, since they are below the desired flight height of the quadrotor. A selection of examples from our dataset is shown in Figure \ref{fig:floorplans}. Among the 50,000 buildings, 45,000 were used for training, 4,900 for validation, and 100 for testing only.

\subsection{Map Predictor}
\label{subsec:predictor}

In contrary to to our previous efforts \cite{9811861, zwecher2020deep}, in this
work we did not assume any prior knowledge of external
walls, to better represent real-world scenarios. In this case, we propose a new DL-based map predictor based on a different network architecture and employs a dynamic thresholding function, aiming to consistently provide accurate predictions to support the exploration planner.

Considering SWaP constraints, we aim to limit the map prediction network to less than 2 TOPs. The resulting map predictor uses a standard Encoder-Decoder network architecture, with 21 encoding layers and 21 decoding layers, with skip connections between each encoding layer and its corresponding decoding layer.

It takes the low-resolution ego-centric 2-D occupancy grid $M_{low}^{ego}$ and generates occupancy probability maps denoted as $M_{pred}^{ego}$. Before feeding the raw predicted map $M_{pred}^{ego}$ into the planner, we apply a dynamic thresholding function to convert the probabilities of all cells $m_{pred}\in M_{pred}^{ego}$ into trinary values $m_{thres} \in \{-1, 0, 1\}$, corresponding to $\{free, unknown, occupied\}$, respectively. The dynamic thresholding function $f_t$ is defined as:
\begin{equation}
\begin{split}
    &  f_t(m_p) = \begin{cases} -1,  & \quad m_p < t_f \\
                             1, & \quad m_p > t_o \\
                             0, & \quad \text{otherwise}
                \end{cases} \\
    & t_f = t_f^{max} \times \min(1,\; 10 \times (\frac{\text{Observed Space}}{\text{Max Floor Plan Size}})^4)  
\end{split}
\end{equation}
where $t_o, t_f^{max}$ are cut-off thresholds for cells being occupied or free, respectively. At the start of exploration, when there is minimal information available, the predictor tends to generate low probabilities for cell occupancy. Consequently, the threshold $t_f$ for cells being considered free is initially set very close to $0$ to ensure conservative predictions. As more regions become observed, this threshold gradually increases until it reaches the maximum cut-off threshold $t_f^{max}$. 

To train the network, we collect partial observations during the first phase of DRL exploration planner training, which is discussed in Sect.~\ref{subsec:training}. Partial observations are treated as training inputs, and each corresponding floor plan is taken as ground truth. The averaged binary cross-entropy loss is used. Empirically, with the dynamic thresholding function and the thresholds we used in Sect.~\ref{subsec:implementation_details}, we achieve $F_1$ score (Eq.~\ref{eq:f1}) greater than $0.92$ when reaching $98\%$ coverage of the entire floor plan in a noise-free simulation environment. This enables the proposed framework to treat the predictor output as observations when computing the space coverage during the exploration. Moreover, with the high $F_1$ score achieved by the prediction module in non-noisy simulation scenarios, it is reasonable to consider the augmented observation map as the result explored map, which further improves the exploration efficiency.

\section{Exploration Planning}
\label{sec:rl_planner}
In this section, we present an exploration planning module that generates exploration actions using a DRL-based planner, converts these actions to waypoints, and plans minimum jerk trajectories toward waypoints.

Different from our previous work~\cite{9811861}, we introduce a new planner with a novel reward function that is associated with the predictor's $F_1$ score. Real-world scenarios inevitably contain noises from various sources, including state estimation, sensor readings, and tracking errors from controllers. To address these challenges, we incorporate realistic sensor models and model the action space using sampled data considering robot dynamics. In order to handle more realistic scenarios and make more elaborate decisions regarding next actions, the proposed planner takes in high-resolution maps as well as low-resolution predictions.

\subsection{Sensor Modeling}
\label{subsec:sensor_modeling}

To enhance the realism of the training simulator and facilitate the sim-to-real transfer, we create a planar sensor model with realistic \gls{fov}, range and sensing noise. For noise modeling, we assign a fixed probability for the adjacent cells to be labeled as occupied, whenever a cell is detected as occupied.

\subsection{Action Space}
To simplify the action space representations, we model all possible actions as a discrete set of six actions, indexed from $0$ to $5$. These actions correspond to hovering, turning right, turning left, moving forward, turning right and moving forward, turning left and moving forward, respectively. However, considering the influence of robot dynamics, the result of an action also depends on the current robot velocities. For example, if the robot receives a yaw-in-place action while moving forward, the actual movement would be a combination of translation and rotation, with translation resulting from deceleration and rotation resulting from the yaw-in-place action. To account for this, we take measurements of robot motions when executing a specific action in the Gazebo simulation. Subsequently, we model the effects of all actions as a function of the previous action, which is then used for simulating robot movement during DRL training.

\subsection{Observation Space}
\begin{figure}[t!]
    \centering
    \includegraphics[width=0.42\textwidth]{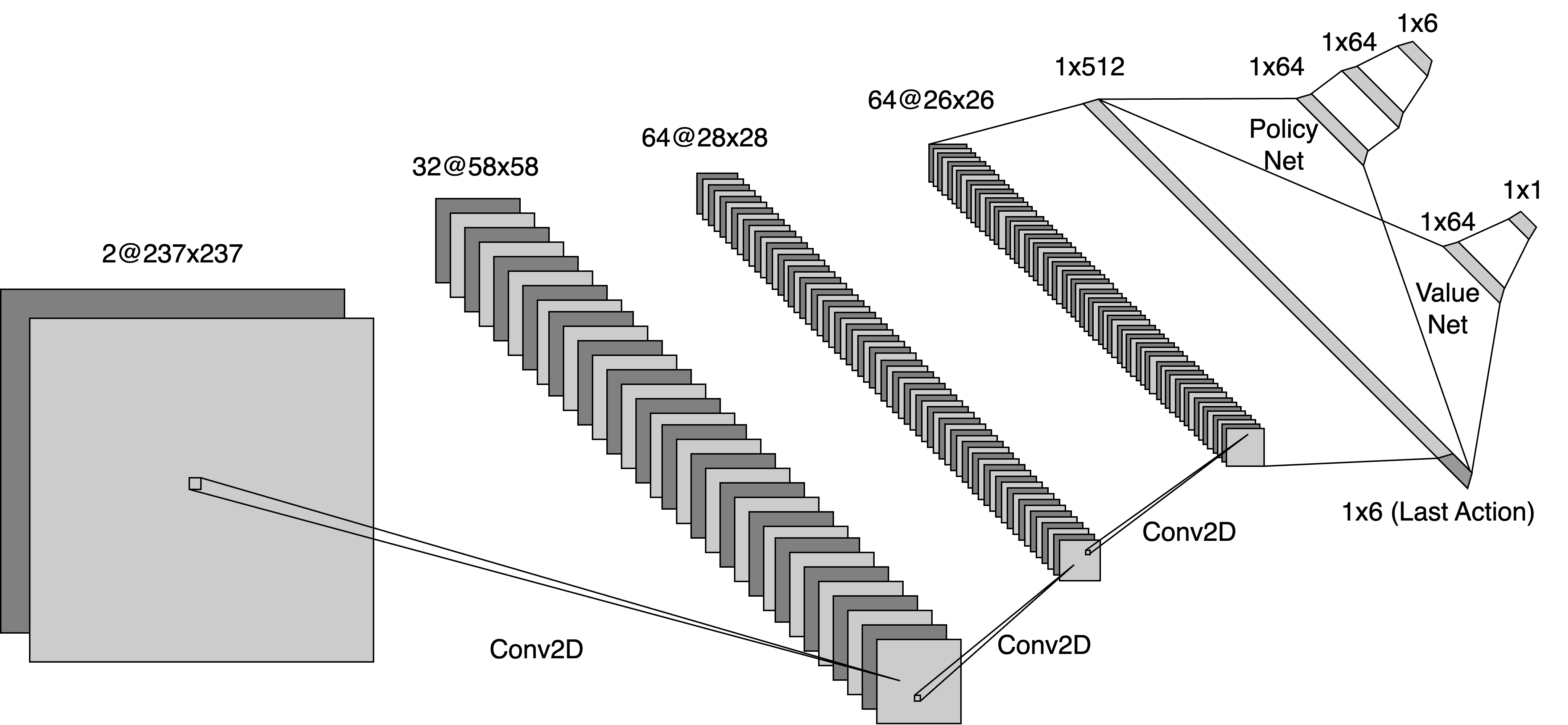}

    \caption{DRL Network Architecture. Features are extracted from the two-layer maps through three standard convolutional layers with ReLU activation. A one-hot vector representing the last action is concatenated with the flattened features. All the features were processed in two different fully connected networks to output the action and the value.}
    \label{fig:rl_nework}
    \vspace{-.7cm}
\end{figure}

The observation space of our framework includes the last action of exploration agent and two-layer maps consisting of ego-centric high-resolution map and ego-centric predicted map. We denote the observation space as $O = \{S_{high}^{ego}, S_{thres}^{ego}, a_{last}\}$. 

While $S_{high}^{ego}$ provides fine-grained environmental details for safe robot navigation in noisy environments, $S_{thres}^{ego}$ enables the agent to select exploration goals that utilize predictions based on prior statistics. Note that both map layers are shifted, padded, rotated, and cropped to keep the robot positioned at the center of the maps with a fixed heading direction. These operations ensure that the observation space changes significantly whenever the robot translates or rotates. We find this feature contributes to a more efficient convergence during training and also saves the need for explicitly recording the agent heading at every step. Furthermore, the two-layer maps are inflated around the occupied cells; in our case, the inflation range is set to approximately match the radius of the robot. The inflated region is referred to as the $non\text{-}flight$ zone, which is classified as a potential collision area and is used in reward calculation as described in Sect.~\ref{subsec:reward}. With the inflation, the two-layer maps now encompass four types of cells: $\{free, unknown, occupied, non\text{-}flight\}$. The size of each map layer is set to a constant value of $237\times237$, ensuring that $S_{thres}^{ego}$ always includes the entire building regardless of the robot's position and heading. The two-layer maps are passed through three standard convolutional-RELU layers and then flattened into a feature vector as shown in Fig.~\ref{fig:rl_nework}. 

The one-hot vector representing the last action is concatenated to the previously mentioned feature vector and fed into both the policy network and the value network. This concatenation is intended to allow the subsequent network layer to learn about the simplified robot dynamics, as indicated by the definition of the action space. As a result, the network learns to rank future actions not only based on environmental maps but also on previous robot motion.

\subsection{Reward Function}\label{subsec:reward}
To guarantee collision-free in the DRL-based planner, considering robot dynamics, we penalize any actions that would lead the agent into potential collision regions in the current or the next time step. The potential collision regions are defined as the union of $non\text{-}flight$ and $occupied$ regions. 
The reward function is defined as:
\begin{equation}
  \mathcal{R}_{a_t}(O_t, O_{t+1}) = -1 + \begin{cases}
     -l,  \quad  \; \text{if collision}\\
     r(a_t), \; \text{otherwise} \end{cases} 
\end{equation}
where $a_t$ represents the action at time $t$. $O_t$ and $O_{t+1}$ denote the observations at $t$ and $t+1$. A constant reward $-1$ is set to encourage the robot to finish the exploration sooner.  Potential collisions are penalized with reward $-l$.
Otherwise, the reward  $r(a_t)$ depends on the changes in the $F_1$ score between the current predicted map after dynamic thresholding, denoted as $M_{thres, t}^{ego}$ at time $t$, and the resulting map, $M_{thres, t+1}^{ego}$ at $t+1$, following the execution of $a_t$:
\begin{equation}
    r(a_t) =  \omega \cdot \left[F_1(M_{thres, t+1}^{ego}) - F_1(M_{thres, t}^{ego})\right]
\end{equation}
where $\omega$ is a constant scaling factor. The $F_1$ score is calculated as:
\begin{equation}
\label{eq:f1}
F_1 = TP / (TP + 0.5(N_M-T))
\end{equation}
where $TP$ is the number of correctly predicted occupied cells, and $N_M$ denotes the total number of interior cells in the floor plan. $T$ counts for the total number of correctly predicted cells including both free and occupied cells within the building's interior.

The dynamic reward $r_d(a_t)$ motivates the robot to observe more structural aspects of the environment rather than exploring free space. While the robot moves toward occupied cells, it in turn benefits the predictor by providing more occupancy information. This design ensures a tight coupling between the \gls{drl}-based exploration planner and \gls{dl}-based map predictor, leading to improved exploration efficiency.

\begin{figure}[t!]
    \centering
    \includegraphics[width=0.9\columnwidth]{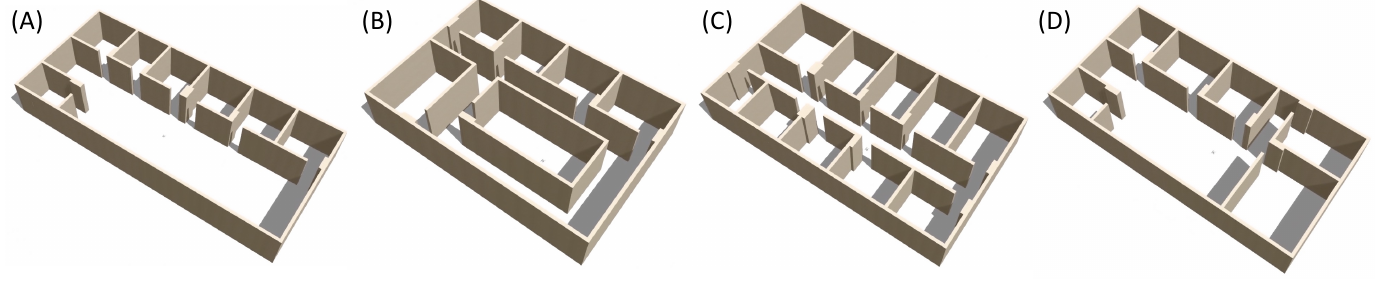}
    \caption{The environments used for benchmarking (A-D) showing in 3-D Gazebo models.}
    \label{fig:benchmark_env}
    \vspace{-.7cm}
\end{figure}
\subsection{Training}\label{subsec:training}
We trained our agent with the PPO algorithm offered by stable baselines 3 using the network described in Fig.\ref{fig:rl_nework}. The training process includes two phases. 

In the first phase, the map predictor is not utilized. Therefore in Eq.~\ref{eq:f1}, term $TP$ becomes the number of observed occupied cells and $N_M - T$ becomes the number of unknown cells within the building's interior. The intention is to allow the agent to first learn the basic exploration strategy, which includes avoiding collisions and prioritizing the exploration of unknown regions. Furthermore, in this phase, the robot's observation at each step is recorded and sampled as inputs for predictor training. In the second phase, building upon the trained policy from phase one, we gradually add predicted information. Specifically, at the beginning of this training phase, the predicted map cutoff thresholds, denoted as $t_f^{max, train}$ and $t_o^{train}$, are set very close to 0 and 1. These thresholds gradually transition to $t_f^{max}$ and $t_o$ as the training proceeds. Note that during each episode, the threshold $t_f$ increases towards $t_f^{max, train}$ according to the dynamic thresholding function.

In each episode, we randomly select a building from the training set. The agent's starting point and heading are sampled uniformly within the building interior. To simulate real-world scenarios, sensor noise is introduced into the training simulator as detailed in Sect.~\ref{subsec:sensor_modeling}.

%% file: tex/Sim_and_Exp.tex
\section{Results and Analysis}
\label{sec:sim_and_exp}

\subsection{Implementation Details}
\label{subsec:implementation_details}
In both our simulation and real-world experiments, the 3-D voxel maps are built and maintained using the OctoMap library \cite{hornung13auro}. The high-resolution is set to $5cm$, while the low-resolution is set to $20cm$. The predictor thresholds were set to $t_f^{max} = 0.04$ and $t_o = 0.94$.

\begin{table*}[!t]
\caption{EXPLORATION STATISTICS FOR FOUR BENCHMARK ENVIRONMENTS IN FIG.~\ref{fig:benchmark_env}}
\scriptsize
\begin{center}
\label{tab:benchmark}
\begin{tabular}{
|P{0.14\textwidth}|
P{0.02\textwidth}P{0.02\textwidth}P{0.02\textwidth}P{0.05\textwidth} | 
P{0.02\textwidth}P{0.02\textwidth}P{0.02\textwidth}P{0.05\textwidth} | 
P{0.02\textwidth}P{0.02\textwidth}P{0.02\textwidth}P{0.05\textwidth} | 
P{0.02\textwidth}P{0.02\textwidth}P{0.02\textwidth}P{0.05\textwidth} | 
}
\hline
& \multicolumn{4}{c|}{\textbf{Building A}} & \multicolumn{4}{c|}{\textbf{Building B}} & \multicolumn{4}{c|}{\textbf{Building C}} & \multicolumn{4}{c|}{\textbf{Building D}} \\ 
\cline{2-17} 
\textbf{Method} & \multicolumn{4}{c|}{\textbf{Path Length (m)}} & \multicolumn{4}{c|}{\textbf{Path Length (m)}} &\multicolumn{4}{c|}{\textbf{Path Length (m)}} & \multicolumn{4}{c|}{\textbf{Path Length (m)}} \\ 
\cline{2-17}
& Min & Max & Avg & Std Dev & Min & Max & Avg & Std Dev & Min & Max & Avg & Std Dev & Min & Max & Avg & Std Dev
\\ 
\cline{1-17}
Frontier\cite{Frontier} & 96.3 & 143.3 & 125.0 & 13.48 & 79.7 & 140.7 & 108.8 & 21.10 & 113.8 & 146.7 & 132.7 & 10.88 & 93.9 & 132.5 & 110.0 & 12.39\\
FUEL\cite{zhou2021fuel}  & 111.1 & 135.7 & 123.6 & 7.74 & 109.2 & 140.3 & 122.0 & 9.41 & 116.7 & 154.4 & 128.6 & 11.72 & 90.1 & 128.0 & 106.0 & 10.17\\
Frontier + Predictor$^{\mathrm{a}}$ & 60.1 & 113.2 & 83.0 & 16.50 & 52.6 & 71.4 & 60.4 & \textbf{4.72} & \textbf{52.6} & 94.6 & 69.2 & 13.90 & \textbf{38.9} & 83.3 & 63.0 & 13.12\\ 
FUEL + Predictor$^{\mathrm{b}}$ & 59.6 & 119.5 & 77.3 & 17.09 & 53.7 & 82.6 & 64.7 & 10.38 & 56.6 & 89.1 & 74.7 & 8.79 & 51.8 & 101.5 & 67.8 & 14.54\\
\rowcolor{lightgray} DRL + Predictor (ours) & \textbf{38.8} & \textbf{58.4} & \textbf{50.2} & \textbf{5.01} & \textbf{40.6} & \textbf{57.9} & \textbf{49.3} & 5.51 & 60.6 & \textbf{74.8} & \textbf{66.3} & \textbf{4.58} & 40.2 & \textbf{64.0} & \textbf{47.4} & \textbf{6.66}\\
\hline
\multicolumn{16}{p{0.8\textwidth}}{$^{\mathrm{a}}$ Nearest Frontier with observed maps fed into predictor, and predictor outputs are treated as explored maps for each time step} \\
\multicolumn{16}{p{0.8\textwidth}}{$^{\mathrm{b}}$ FUEL with observed maps fed into predictor, and predictor outputs are treated as explored maps for each time step}
\end{tabular}
\vspace{-.7cm}
\end{center}
\end{table*}

\begin{figure}[t!]
    \centering
    \includegraphics[width=1.0\columnwidth]{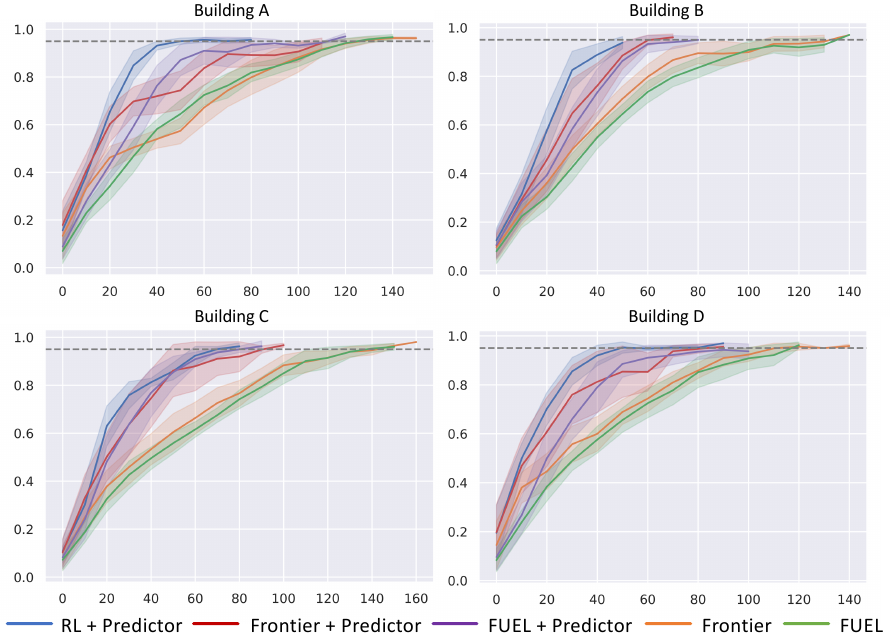}
    \caption{Results with the benchmark environments: The X-axis represents trajectory length in meters. The Y-axis represents room coverage, which is calculated by dividing the number of explored cells by the total number of cells. For methods with predictors, free or occupied cells in predictor outputs are counted as explored. For methods without predictors, free or occupied cells in observed maps are counted as explored. The horizontal dotted line signifies 95\% coverage. Solid lines illustrate the average statistics derived from 10 experiments for each environment, while the translucent regions surrounding the solid lines depict corresponding standard deviations.}
    \label{fig:benchmark_stats}
    \vspace{-.7cm}
\end{figure}

\subsection{Simulation Experiments}

To demonstrate the effectiveness and generalizability of our proposed framework, we conducted extensive evaluations in Gazebo simulations, where robot dynamics, sensor noises, and real-world physics are well-modeled. We benchmarked the proposed method against the state-of-the-art as well as the classical exploration algorithm using four buildings from the test dataset, each possessing distinct topological characteristics as shown in Fig.~\ref{fig:benchmark_env}. Within each building, $10$ experiments are carried out, all initiated from the same starting position. For each trial, we recorded the path length at the point of achieving 95\% coverage of the total space. Subsequently, we computed the relevant statistics, which are presented in Tab.~\ref{tab:benchmark}. Furthermore, the ratios between space coverage and path length are visualized in Fig.~\ref{fig:benchmark_stats}. The proximity of the curved lines to the upper-left corner signifies the higher efficiency of the respective methods. In all four benchmarking environments, the proposed framework achieves the best performance. The results show that our method outperforms the classic and state-of-the-art methods by $50$ to $60\%$.

Notice that we introduced 'Nearest Frontier + Predictor' and 'FUEL~\cite{zhou2021fuel} + Predictor' methods, both of which incorporate an additional inference step subsequent to the base method. In these approaches, the agent still explores the environment following the foundational method, which relies on the observed map. Then, at each discrete time step, the observed map is fed into the map predictor, and space coverage is calculated utilizing the predicted map. Comparisons of our proposed framework with these two methods reveal that the superiority of our approach doesn't solely come from the map predictor. Instead, the DRL planner is tightly coupled with the predictor. The planner learns to guide the agent to traverse regions that provide the map predictor with sufficient cues to predict the entire map, thus enabling the robot to finish explorations with shorter trajectories.

\subsection{Real-world Experiments}
\subsubsection{Computational Requirements}
\begin{figure}[t!]
    \centering
    \includegraphics[width=1.0\columnwidth]{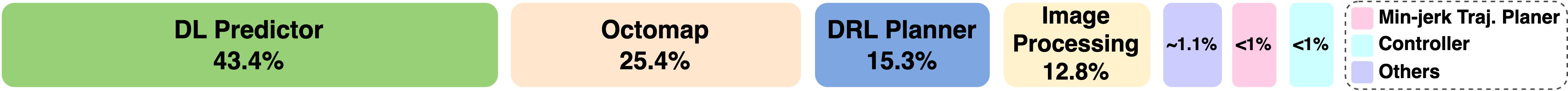}
    \caption{CPU utilization of the entire software stack. The entire software stack uses $32.7\%$ of the onboard CPU, which is further broken into seven blocks. `Others' includes the utilization of other nodes in ROS.}
    \label{fig:cpu}
    \vspace{-.7cm}
\end{figure}

Firstly, we quantify the CPU utilization of the full autonomy stack on the platform's onboard i7-10710 CPU, as shown in Fig.~\ref{fig:cpu}. The DRL planner's inference takes $\sim8ms$, and the predictor's inference takes $\sim135ms$ with the onboard CPU.

To ensure that the system is able to be deployed on other \gls{swap} constrained platforms, we tested both trained network on the $4$ TOPS Coral TPU~\cite{coral}. The DRL planner's inference takes $\sim8ms$ and the predictor's inference takes $\sim150ms$. The low inference times observed suggest that the system remains viable for deployment on \gls{swap}-constrained platforms equipped with specialized neural processors. Moreover, we have successfully tested the ability to run the remainder of the software stack on Qualcomm Snapdragon 821 ARM SoC.

\subsubsection{Autonomous Exploration Experiments}
\begin{figure}[t!]
    \centering
    \includegraphics[width=0.93\columnwidth]{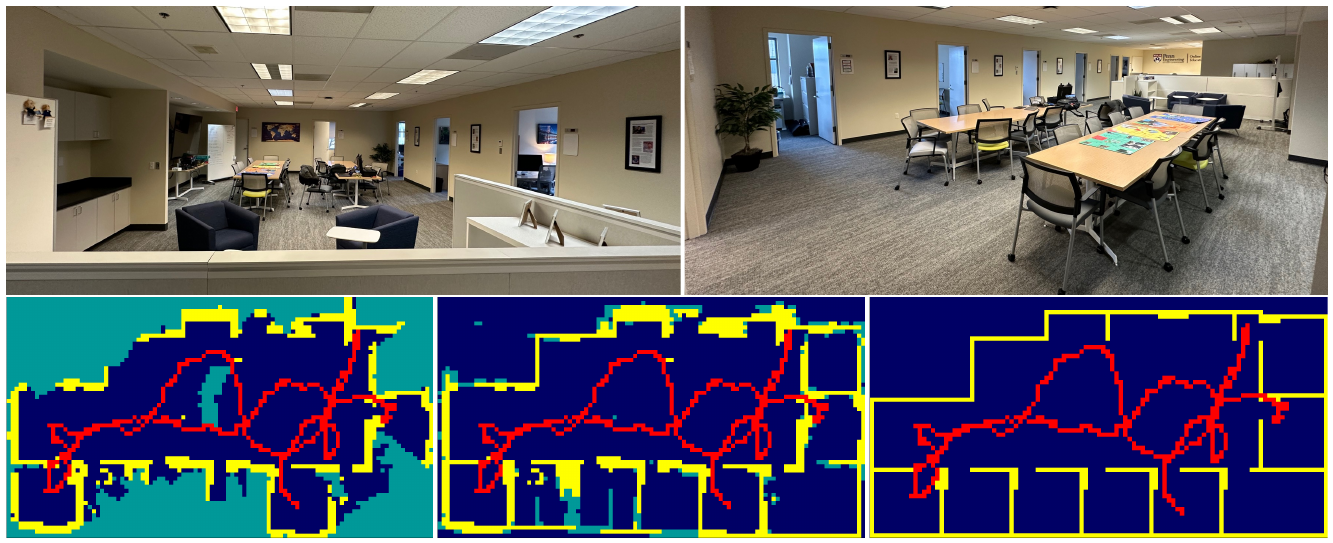}
    \caption{Representative results of experimentation with the Falcon 250 in an office suite of roughly $10m \times 20m \times 3m$ dimensions, comprising eight office rooms. On the bottom panel, the right figure shows the ground truth occupancy map that we constructed based on the floor plan. The left figure shows the final observed 2-D occupancy map. The middle figure shows the final predicted map. Free, unknown, and occupied cells are marked as dark blue, light blue, and yellow colors, respectively.}
    \label{fig:realworld}
    \vspace{-.7cm}
\end{figure}

For real-world experiments, Falcon 250 platform (Fig.~\ref{fig:platform}) was used and the odometry data from the VOXL VIO system\cite{voxl} was utilized. The entire software stack including both network models was run onboard.
Our real-world experiments were conducted within an office suite of roughly $10m \times 20m \times 3m$ dimensions. This space comprises a common area and eight rooms of varying sizes, as illustrated in Fig.~\ref{fig:realworld}. Notice that when projecting 3-D voxel maps onto 2-D occupancy grid maps, voxels below a height of 1.5 meters are ignored. We set the desired flight height of the quadrotor to be 1.7 meters. This means that the DRL observation space is only aware of high obstacles like walls and cabinets, while ignoring low objects such as chairs and tables.

The MAV initiated its flight within the upper-right office. After scanning the majority of the initial room, the agent is able to confidently predict the rest of it. Hence, the agent departed and explored the hallway in a right-to-left manner. Subsequently, it navigated into the lower-left room, which provided the predictor with the greatest amount of information about the building's perimeter. Upon exiting the aforementioned room and scanning the upper part of the building, the drone entered a room situated in the middle of the lower portion, which provides the predictor with sufficient clues regarding the layout of offices in the lower segment of the building.

Compared to conventional exploration methods, the DRL planner is designed to explore regions that are likely to provide crucial cues for the predictor to deduce building structure. The MAV's behaviors throughout the experiment demonstrate that the DRL planner and the predictor are tightly coupled. Every action decided by the DRL planner is aligned with the objective of helping map predictor generate comprehensive and precise predictions, which is consistent with the rationale of our customized reward function.

%% file: tex/Conclusions.tex
\section{Conclusion}
\label{sec:conclusion} 
In this paper, we presented a framework that tightly couples deep learning and reinforcement learning for autonomous micro aerial vehicles exploring unknown indoor environments. Our proposed framework contains a mapping and prediction module, which generates high-accuracy predictions of the environment based on the observed map, and an exploration planning module, which learns to leverage the predictions for efficient exploration enabling better predictions.  Throughout our work, we accounted for real-world sensor noise and robot dynamics, which contribute to a smoother sim-to-real transfer process. Extensive testing in the Gazebo simulation and benchmarking against classical and state-of-the-art methods demonstrate the effectiveness of our framework. Successful experiments illustrate that the sensing and computation requirements are compatible with a SWaP-constrained MAV platform. A natural question is if the map predictor and exploration module can be extended in three-dimensions. This and the extension to multiple robots are directions for future research. While our experiments show that the entire software stack can be easily run on an Intel i7-10710U CPU, it is worth noting that the VIO-based state estimation algorithm runs on an independent VOXL board~\cite{voxl}. Similarly it is possible to run the deep learning inference engine on a specialized TPU like the Coral~\cite{coral} processor. There is clearly an opportunity to optimize the hardware architecture to support light weight computation for the software stack described in the paper and this is also a direction for future research.